\documentclass[twocolumn]{article}
\usepackage{paralist}
\usepackage{enumitem}
\usepackage{graphicx}
\usepackage{upgreek,textgreek}
\usepackage{amsmath}
\usepackage{amssymb}
\usepackage{bm}
\usepackage{yhmath}
\usepackage{mathtools}
\usepackage{array,multirow}
\usepackage{algorithmic}
\usepackage{paralist}
\usepackage{enumitem}
\usepackage{graphicx}
\usepackage{upgreek,textgreek}
\usepackage{bm}
\usepackage{yhmath}
\usepackage{mathtools}
\usepackage{array,multirow}
\usepackage{algorithmic}
\usepackage{balance}
\usepackage{booktabs}
\usepackage{subcaption}
\usepackage{xcolor}
\usepackage{url}
\usepackage{hyperref}

\DeclarePairedDelimiter{\ceil}{\lceil}{\rceil}
\begin{document}
\date{}

\title{DietCNN: Multiplication-free Inference for Quantized CNNs\\
{\large ({\color{blue} Supplementary Material)}}}

\author{
Swarnava Dey \\
\textit{TCS, Research \& IIT Kharagpur} \\
\textit{Tata Consultancy Services Ltd.}\\
Kolkata, India \\
\texttt{\small swarnava.dey@tcs.com}
\and
Pallab Dasgupta \\
\textit{Computer Science \& Engineering} \\
\textit{Indian Institute of Technology Kharagpur}\\
Kharagpur, India \\
\texttt{\small pallab@cse.iitkgp.ac.in}
\and
Partha P Chakrabarti \\
\textit{Computer Science \& Engineering} \\
\textit{Indian Institute of Technology Kharagpur}\\
Kharagpur, India \\
\texttt{\small ppchak@cse.iitkgp.ac.in}
}

\maketitle
\begin{abstract}
\textit{The rising demand for networked embedded systems with machine intelligence has been a catalyst for sustained attempts by the research community to implement Convolutional Neural Networks (CNN) based inferencing on embedded resource-limited devices. Redesigning a CNN by removing costly multiplication operations has already shown promising results in terms of reducing inference energy usage. This paper proposes a new method for replacing multiplications in a CNN by table look-ups. Unlike existing methods that completely modify the CNN operations, the proposed methodology preserves the semantics of the major CNN operations. Conforming to the existing mechanism of the CNN layer operations ensures that the reliability of a standard CNN is preserved. It is shown that the proposed multiplication-free CNN, based on a single activation codebook,  can achieve  4.7x, 5.6x, and 3.5x reduction in energy per inference in an FPGA implementation of MNIST-LeNet-5, CIFAR10-VGG-11, and Tiny ImageNet-ResNet-18 respectively. Our results show that the DietCNN approach significantly improves the resource consumption and latency of deep inference for smaller models, often used in embedded systems. Our code is available at:} \url{https://github.com/swadeykgp/DietCNN}
\end{abstract}
\\The paper is available at:
\begin{verbatim}
    S. Dey, P. Dasgupta and P. P. 
    Chakrabarti, "DietCNN: 
    Multiplication-free Inference
    for Quantized CNNs," 2023 
    International Joint 
    Conference on Neural Networks
    (IJCNN), Gold Coast, 
    Australia, 2023, pp. 1-8, 
    doi: 10.1109/
    IJCNN54540.2023.10191771.
\end{verbatim}
\newcommand{\vect}[1]{\bm{#1}}

{\small
\definecolor{Gray}{gray}{0.5}
\renewcommand\UrlFont{\color{Gray}\rmfamily}
\bibliographystyle{ieee_fullname}

}

\appendix
\section{Summary of Appendices}
The supplementary material is organized as follows:
\begin{itemize}
    \item \ref{sec:dics_others_supp} presents the transformers for layers other than the convolutional layer. This part of the appendix supplements Sec.III-B of the main paper.
    \item \ref{sec:dietben} presents a detailed analysis of the theoretical benefits of DietCNN transformation. This part of the appendix supplements Sec.III-B of the main paper.
    \item \ref{sec:assocadd} presents the experiments done on DietCNN symbolic addition associativity. This part of the appendix supplements Sec.III-D of the main paper.
    \item \ref{sec:retraining} presents the second mode of training for generating a DietCNN model from the scratch. This part of the appendix supplements Sec.III-D of the main paper.
    \item \ref{sec:fpga} presents the methodology followed for measuring the energy of DietCNN inference. Results presented in Table 1 of the main paper are based on this methodology. This part of the appendix supplements Sec.IV of the main paper. This section also explains the calculation of the number of MACs / Lookups presented in Table 1 of the main paper.
    \item \ref{sec:eval_supp} presents the detailed analysis of the memory requirement and overhead for the DietCNN models. It also presents some ablation studies on the most important hyperparameters of DietCNN. This part of the appendix supplements Sec.IV of the main paper.
    \item \ref{sec:assets} presents the details about all the building blocks used by us in this work with their licensing details.
\end{itemize}

\section{Discrete Transformers for Other Layer Operations}
\label{sec:dics_others_supp}
This part of the appendix supplements Sec.III-B of the main paper.

\subsection{Activation Functions}
\label{sec:actfun}
\begin{figure*}[htb]
  \centering
   \includegraphics[width=0.6\linewidth]{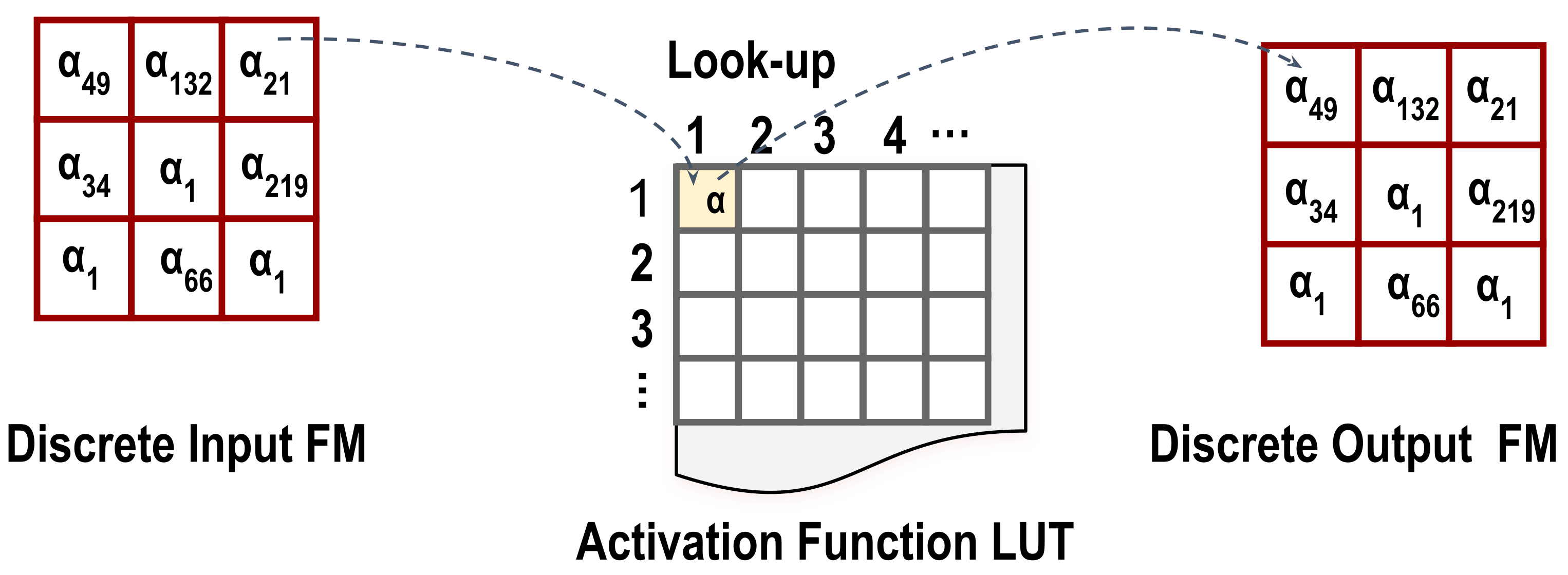}
   \caption{The discrete transformer for an activation layer function, e.g., ReLU, sigmoid.}
   \label{fig:relu_layer}
\end{figure*}

In the DietCNN design, an activation function can be computed by a symbol-wise look-up, during the CNN inference phase. Once the activation LUT for activation is pre-computed (see Sec. 3.1.2 of the main paper), we need one look-up per symbol, for each symbol of the input symbolic feature map. This is depicted in Fig.~\ref{fig:relu_layer}.
This method is fast and does not suffer any accuracy drop when we replace all ReLU activations with sigmoid in the MNIST LeNet-5~\cite{lecunmnist}.

\subsection{Fully Connected Layer}
\label{sec:fcfun}
\begin{figure*}[htb]
  \centering
   \includegraphics[width=0.98\linewidth]{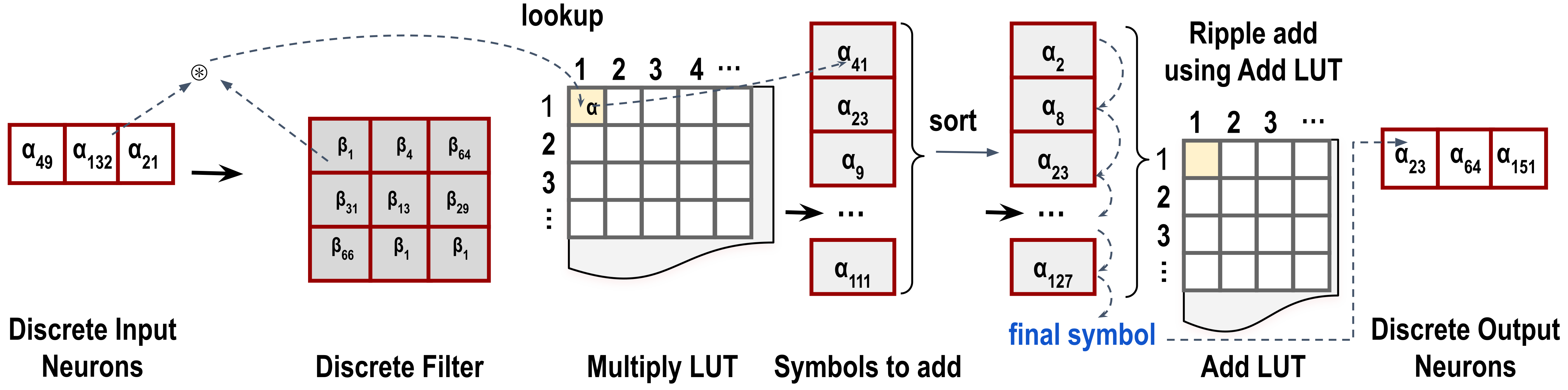}
   \caption{The multiplication-free discrete transformer for fully connected layer}
   \label{fig:fc_layer}
\end{figure*}

A DietCNN fully connected layer is implemented in the same way as a convolutional layer, as shown in Fig.~\ref{fig:fc_layer}.

\section{Retraining for Recovering Accuracy}
\label{sec:retraining}
This part of the appendix supplements Sec.III-D of the main paper. This method can be used if the recommended methodology (Sec.III-D of the main paper) fails for some network/ dataset.

Let us consider that a set of images $x_1, x_2,\ldots,x_N$, $x_i \in \mathbb{R}^{H \times W \times C}$ and a set of labels $y_1, y_2,\ldots,y_k$, $y_i \in \mathbb{R}$, constitute the original training dataset {\cal F}, of pre-trained CNN $\mathcal{N}_\theta$. The CNN parameters $\theta$ include weights $\vect{w}$ and biases $\vect{b}$. We aim to re-train and recover the accuracy of the target DietCNN  $\mathcal{N}^{Diet}_{\widehat{\theta}}$, where the $\widehat{\theta}$ include the symbolic weights $\vect{\hat{w}}$ and biases $\vect{\hat{b}}$. The forward pass and symbolic parameters of $\mathcal{N}^{Diet}_{\widehat{\theta}}$ can be derived from  $\mathcal{N}_\theta$, using the DietCNN transformation.
 \newcommand{\viss}[1]{\mathbf{V}}
 \newcommand{\vis}[1]{\mathbf{v}_{#1}}
 \newcommand{\prd}[2]{\hat{y}^{#2}_{#1}}
 \newcommand{\gt}[2]{y^{#2}_{#1}}
 \newcommand{\raw}[2]{\mathbf{I}^{#1}_{#2}}
 \newcommand{\prmm}[1]{\mathbf{W}_{#1}}
 \newcommand{\cnnw}[1]{\mathbf{w}^{#1}}
 \newcommand{\prms}[1]{\mathbf{\ww}_{#1}}
 \newcommand{\reg}[1]{\mathcal{R}(#1)}
 \newcommand{\task}{t}
 \newcommand{\losst}[3]{L_{#1}\left(#2,#3\right)}
  \newcommand{\losstwo}[2]{L_{#1}\left(#2\right)}
  \newcommand{\lossd}[3]{L'_{#1}\left(#2,#3\right)}
\newcommand{\lossdd}[4]{\nabla_{#4}L_{#1}\left(#2,#3\right)}
 \newcommand{\mydelta}[2]{\delta_{#1,#2}}
 \newcommand{\cnn}{0}

\newcommand{\bi}{i}
\newcommand{\bs}{B}
\newcommand{\mb}{\mathcal{B}}
\newcommand{\tast}{p}

\newcommand{\grad}[1]{{\mathrm{\mathbf{d}}#1}}
\newcommand{\cnt}[1]{\mathrm{\mathbf{c}}_{#1}}
\newcommand{\setto}{\leftarrow}
 \begin{table}[!t]
   \centering
   \scalebox{0.96}{
   \parbox{.7\linewidth}{       
 	\hrule
 	\vspace{3pt}
 	Algorithm for Re-training DietCNN
 	\hrule
 	\begin{algorithmic}
 		\FOR{$m = 1$ to $M$}   
 		\STATE \COMMENT{sample minibatch}
 		\STATE  $\mb \setto \{x_1,\ldots,x_{\bs}\}$ with $x_i \sim U[1,N]$ 
 		\STATE \COMMENT{initialize minibatch gradient accumulators}
 		\STATE $\grad{\vect{w}} \setto 0$ 
 		\STATE $\grad{\vect{b}} \setto 0$ 
 		\FOR{$\bi \in \mb$}
 		\STATE \COMMENT{gradient computation with DietCNN}
 		\STATE $ \hat{y} \setto  \mathcal{N}^{Diet}_{\widehat{\theta}}(x_i,\vect{\hat{w}},\vect{\hat{b}}) $
 		\STATE $ \grad{\vect{w}} \setto \lossdd{}{\prd{}{}}{\gt{}{\bi}}{\vect{w}}$
 		\STATE $ \grad{\vect{b}} \setto \lossdd{}{\prd{}{}}{\gt{}{\bi}}{\vect{b}}$
 		\ENDFOR
 		\STATE \COMMENT{updating parameters of standard CNN}
 		\STATE $\vect{w} \setto \vect{a} - \lambda\frac{1}{\bs}\grad{\vect{a}}$
 		\STATE $\vect{b} \setto \vect{b} - \lambda\frac{1}{\bs}\grad{\vect{b}}$
 		\STATE \COMMENT{re-create symbolic weights $\vect{\hat{w}}$, and LUTs $\nu^m$, $\nu^a$,$\nu^n$,$\nu^b$ for DietCNN (See Sec. 2.1.1 \& 2.1.2 of main paper)}
 		\ENDFOR
 		\vspace{.2cm}
 		\hrule
 	\end{algorithmic}
 	\caption{\small Pseudocode for the proposed minibatch stochastic gradient descent algorithm for back-propagation training of DietCNN parameters. 
    }	
 	\label{tab:sgd}}}
 \end{table}


As shown in the training algorithm (Tab.~\ref{tab:sgd}), the loss is calculated using the forward pass of the transformed DietCNN version $\mathcal{N}^{Diet}_{\widehat{\theta}}$, and this loss is backpropagated for each minibatch on the weights and biases of the original CNN.

At the end of each minibatch, the symbolic weights and LUTs are generated from these updated weights and biases. To generate the symbolic weights, the weights and biases of $\mathcal{N}_\theta$ are clustered, represented as a codebook (See Sec. 3.1.1 \& Sec. 3.1.2 of the main paper). The weights are clustered using the K-means++ algorithm~\cite{arthur2007kmeansp} of the Scikit-learn library~\cite{scikit-learn}. The weight clustering and LUT generation, for $64$ centroids, takes approximately $10$ seconds in our training setup.

\par It may be noted that the Image (with intermediate feature maps) codebook needs to be prepared only once for a given dataset. We perform the clustering for that with a fast clustering library, called FAISS~\cite{johnson2021faiss}. It takes approximately $10$ minutes to build a $128$ centroid clustering index for MNIST. For the complete ImageNet dataset and the intermediate feature maps generated by AlexNet, approximately $36$ hours is needed to build a faiss clustering index. The use of a single codebook for images and intermediate feature maps, resulting in the flow of a restricted set of symbols throughout the network, differentiates DietCNN from all earlier works on quantized DNN inference~\cite{quaint,jacob2018quaint,ZhangYangYeECCV2018,cnngolami,chen2021aqd}.   

The training algorithm shown in Tab.~\ref{tab:sgd}, is implemented in the PyTorch framework. The random seed for all the libraries is set to $0$. We train LeNet-5 on the MNIST dataset, using Stochastic gradient descent (SGD) with Negative Log-Likelihood loss for $60$ epochs. For the DietCNN variant of LeNet-5, we try both the post-facto, training-free method and the fine-tuning method that takes $3$ epochs to recover the 6\% accuracy drop.  For VGG-11 on CIFAR-10, we use Cross Entropy loss with SGD. The starting learning rate is $0.1$, with a Cosine Annealing learning rate scheduler. We train the model for 200 epochs to reach 91.9\% accuracy. For the DietCNN variant of VGG-11, we try the post-facto, training-free method and reach 89.6\% accuracy.

\section{Benefits of Discrete Convolution Transformer}
\label{sec:dietben}
This part of the appendix supplements Sec.III-B of the main paper. We reproduce the equations for the standard convolution and the discrete transformer for that here for easy reference:

The standard CNN is a convolutional layer operation is stated in Eqn.~\ref{eq:eq_conv}.
\begin{equation}
 M_{ijn} = \sum_{m=0}^{c-1}\sum_{i=0}^{h-k}\sum_{j=0}^{w-k} (x_{(i:i+k)(j:j+k)(m)} \circ f_{(1:k)(1:k)mn}),
\label{eq:eq_conv_supp}
\end{equation}
where, for any input feature map $x \in \mathbb{R}^{h \times w \times c}$, $x_{(p:p+r)(q:q+r)}$ denotes the $r \times r$ 2D convolution window with the left-top vertex and right-bottom vertex at the locations ($p$, $q$) and ($p+r$, $q+r$). 

The DietCNN transformed CNN convolutional layer operation is stated in Eqn.~\ref{eq:eq_convsym}.
\begin{multline}
 M_{ijn} = \\ \mid \mathit{f^{add}}( \{\nu^m( \tau(x_{(i:i+k)(j:j+k)(m)}),\tau(f_{(1:k)(1:k)mn})) \\ \mid 0 \leq m \leq c, 0 \leq i \leq h-k+1, 0 \leq j \leq w-k+1 \}) 
\label{eq:eq_convsym_supp}
\end{multline}
where the mapping $\mathit{f^{add}}: (\alpha_{i_1}, \alpha_{i_2}, \ldots,\alpha_{i_n})  \rightarrow \alpha_k$, takes a bag of symbols and adds the symbols one by one, accumulating the result, using $\nu^{a}$. Note that, $M_{ijn}$ is also in the symbol domain. 

Considering Eqn.~\ref{eq:eq_conv_supp}, we need to perform $ c \times k \times k$ MAC operations to compute one scalar value at location $(i,j)$ of an output feature map. With $n$ output filters the total number of MAC operations requires is, $L^c_{ops}$, which can be expressed as follows:
\begin{equation}
L^c_{ops} = c \times k^2 \times n \times (h - k + 1)^2,
\label{eq_stdmac}
\end{equation}
where the input feature map is from an input space in $\mathbb{R}^{h \times w \times c}$ and each output feature map has a resolution of $(h - k + 1) \times (h - k + 1)$.

With discrete convolution in Eqn.~\ref{eq:eq_convsym_supp}, we need to perform exactly the same number of look-ups for multiplication and addition, as we do not change the semantics of the operation. This is true if we use a $1\times1$ patch size for a symbol. In a more general form, the total number of look-ups and {\em add} operations required to derive an output symbol at location $(i,j)$ of the $n^{th}$ output filter, can be expressed as follows:
\begin{equation}
L^{\widehat{c}}_{ops} = c \times \ceil[\big]{(k/P)^2} \times n \times (h - k + 1)^2,
\label{eq_dietop}
\end{equation}
where $L^{\widehat{c}}$ is the discrete convolutional layer, corresponding to the standard convolutional layer $L^c_{ops}$.  
Let a MAC operation, ${\cal M}$, multiply LUT look-up and add LUT look-up take $M$, $L$ and $A$ units of time respectively. The speedup factor can be expressed as follows:
\begin{equation}
\mathbf{\Delta^s} = \left( \frac{L^c_{ops} \times M}{L^{\widehat{c}}_{ops} \times  (L + A)} \right) \\
= \left( \frac{P^2 \times M}{L + A} \right).
\label{eq_symspeed}
\end{equation}

The value of $\mathbf{\Delta^s}$ becomes $\left( \frac{M}{L + A} \right)$ when we discretize at a pixel level, i.e, $P = 1$.



\section{Experiments on Associativity of Symbolic Addition}
\label{sec:assocadd}
This part of the appendix supplements Sec.III-D of the main paper.

We simulate symbolic addition for a set of symbols, out of all the symbols in the image codebook. We choose the number of symbols to add based on the number of symbols generated at different layers of the MNIST LeNet~\cite{lecunmnist} and AlexNet~\cite{alexnet} architecture.
The experiment is conducted in the following manner:
\begin{enumerate}
    \item Number of symbols to be added decided.
    \item An initial random sequence is generated for the number of symbols to be added.
    \item The expected sum is calculated corresponding to the values (centroid vector) of the initial symbols sequence, and then this sum is converted to a symbol using the image dictionary and codebook.
    \item $1000$ different permutations are generated for the same set of symbols.
    \item Each of the randomly generated sequence of symbols is {\em symbolically added} and the resulting symbol is compared to the expected symbol generated in step 3
\end{enumerate}

The result of the above comparison results in three cases, where the symbols are either the same, or near to each other, or far from each other. The {\em nearness} is defined by the clustering distance between the centroids. For instance, in a dictionary with $128$ symbols (centroids), an index search with a given patch, returns all the centroids nearest to that patch, in order of distance. We find that using any one of the five closest centroids for discretization, instead of the closest one, does not result in an accuracy drop. Based on this empirical evidence, consider the symbolic addition result {\em near}, if it is within $5$ symbols from the expected sum, in terms of clustering distance.   
\begin{table}[!t]
\caption{\small Experiments of Associativity of Symbolic Addition for $ 1 \times 1$ patches. {\em Near} denotes the symbols which are within the $5$ nearest centroids to the expected sum, in terms of the clustering distance. The number of symbols to be added (column 2) represents the symbols generated at the network layers specified in column 3. The last three columns add to $1000$, that is, the number of different orders in which the symbols were added. } 
\centering
\scalebox{0.85}{
    \begin{tabular}{ccccc}
        \toprule[.4mm]
        \textbf{\# Symbols} & \textbf{Simulates} & \textbf{Same} & \textbf{Near} & \textbf{far} \\
        \midrule[0.4mm]
        840 & LeNet FC3 & 253 & 342 & 405  \\
        \midrule[0.1mm]
        10080 & LeNet FC2 & 239 & 352 & 409  \\
        \midrule[0.1mm]
        48000 & LeNet FC1 & 259 & 356 & 385  \\
        \midrule[0.1mm]
        12288 & AlexNet C1 & 267 & 354 & 379  \\
        \midrule[0.1mm]
        73728 & AlexNet C2 & 260 & 321 & 419  \\
        \midrule[0.1mm]
        98304 & AlexNet C3 & 264 & 351 & 385  \\
        \bottomrule[0.4mm]
        \end{tabular}
    }
\label{tab:associativity}
\end{table}
	\begin{figure*}[htbp]
		\begin{tabular}{m{0.45\hsize}m{0.45\hsize}}
			\begin{subfigure}[t]{0.4\textwidth}
				\centering
				\includegraphics[width=0.95\textwidth]{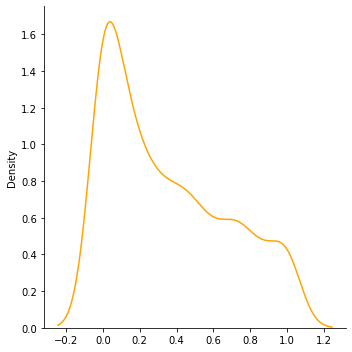}
				\caption{}\label{fig:orig2}
			\end{subfigure}
			
			\begin{subfigure}[t]{0.4\textwidth}
				\centering
				\includegraphics[width=0.95\textwidth]{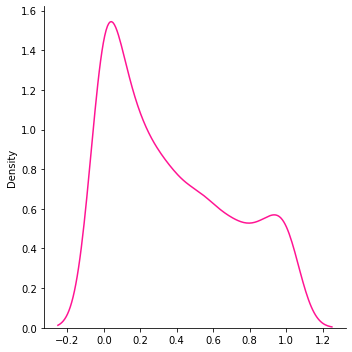}
				\caption{}\label{fig:unclear1}
			\end{subfigure}
			&
			\begin{subfigure}[t]{0.4\textwidth}
				\centering
				\includegraphics[width=0.95\textwidth]{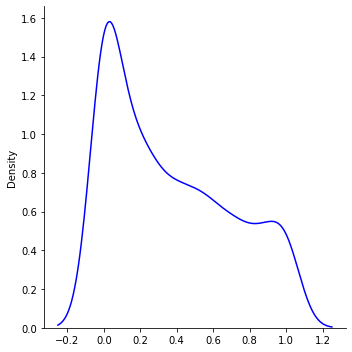}
				\caption{}\label{fig:unclear2}
			\end{subfigure}
			
			\begin{subfigure}[t]{0.4\textwidth}
				\centering
				\includegraphics[width=0.95\textwidth]{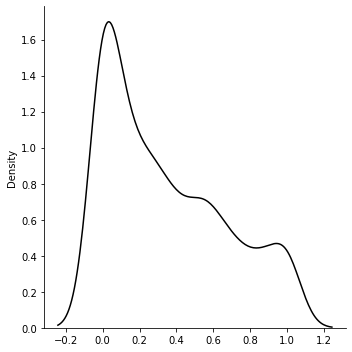}
				\caption{}\label{fig:clear1}
			\end{subfigure}

		\end{tabular}
		\caption{Distribution Visualization: x-axis L2 distance from expected sum(centroid), y-axis frequency of symbols, each plot shows a simulation of the symbol addition experiments considering the input and output channels of some standard CNNs:
			\subref{fig:orig2} Input: 10, Output: 84;
			\subref{fig:unclear1} Input: 120, Output: 84;
			\subref{fig:unclear2} Input: 192, Output: 384;
			\subref{fig:clear1} Input: 384, Output: 256;
		}\label{fig:feature-maps}
	\end{figure*}

The above experiments clearly explain the accuracy drop while adding a large number of symbols. However, the workaround of adding symbols in a particular order helps in recovering the accuracy with retraining.  

As stated in the main paper, we use a patch size of $1$, that is we discretize at a pixel level, for all experiments. This ensures that we can use the DietCNN inference methodology on top of standard quantization. In that case, we need to replace the clustering-based methodology, with a quantization-based approach to create the alphabets that contain the symbols to represent activations and filters. However, the other reason for using clustering instead of quantization is for the generalization of the symbols. If we use larger patches as symbols in the future, we can use some of the well-established image similarity search measures to group the image patches.

\section{Details of DietCNN Memory Overhead}
\label{sec:eval_supp}
This part of the appendix supplements Sec.IV of the main paper.

The DietCNN inference is designed to scale arbitrarily without accuracy degradation. In practice, its accuracy depends on the number of symbols used to represent the input, intermediate feature maps, and filters. 
We have 3 hyperparameters for symbols. Number of symbols for input and activations (N\_CLUSTERS). There are two more, namely, the number of symbols representing all convolutional layer filters (N\_CFILTERS) and the number of symbols representing all fully connected layer filters (N\_FFILTERS).

For N\_CLUSTERS, we have got strong evidence for taking 512 as the upper bound. The experiments in Tab.~\ref{tab:imagenetcleanacc} demonstrates that if we pass a discretized image from the ImageNet dataset to a CNN, the inference accuracy remains within 0.4\% if 2048 symbols are used, and within 1\% if 512 symbols are used.

The experimental setup corresponding to the results in Tab.~\ref{tab:imagenetcleanacc} is as follows:

In Sec. 3.1.1 of the main paper we define a mapping $\tau$ to convert images, and feature maps to their symbolically coded counterparts. We have also implemented a reverse mapping  $\tau_R$, which reconstructs an image from the symbolically coded counterpart. The composite function, $\tau_R(\tau(x))$, where $x$ is an image (or a feature map or a filter), extracts patches from an image (as described in Sec. 3.1.1 of the main paper), replaces those with a representative centroid patches and reconstructs the image. This reconstructed image is not an exact copy of the original image but an approximation created with centroid patches. 

The CNN models referred to in Tab.~\ref{tab:imagenetcleanacc} are PyTorch models with pre-trained weights (\url{https://pytorch.org/vision/stable/models.html}). Please note that we have not implemented the DietCNN transformation for these models yet. We pass the reconstructed approximation of the original image to these networks to evaluate the inference accuracy:  

\texttt{Image(i) -> $\tau_R\tau(i)$ -> Standard CNN -> result}

Although DietCNN operations are pre-computed on full precision to prevent accuracy drop, the symbolic coding  $\tau$ is an approximation that can have an effect on the CNN inference. We wanted to empirically evaluate the extent to which accuracy drops due to this transformation on larger datasets.  We find that the accuracy drop is negligible. Along with this, we also found that the accuracy of the LUT-based inference is primarily dependent on this dictionary of input and activations (See Tab.~\ref{tab:hyper}.

\begin{table}[ht]
\caption{Experimental results show that if we pass a discretized image from the ImageNet dataset to a CNN, the maximum inference accuracy drop is 1\% if 512 symbols are used. Table reproduced from SymDNN~\cite{swasym}.} 
\centering
\scalebox{0.6}{
\includegraphics[width=0.8\textwidth]{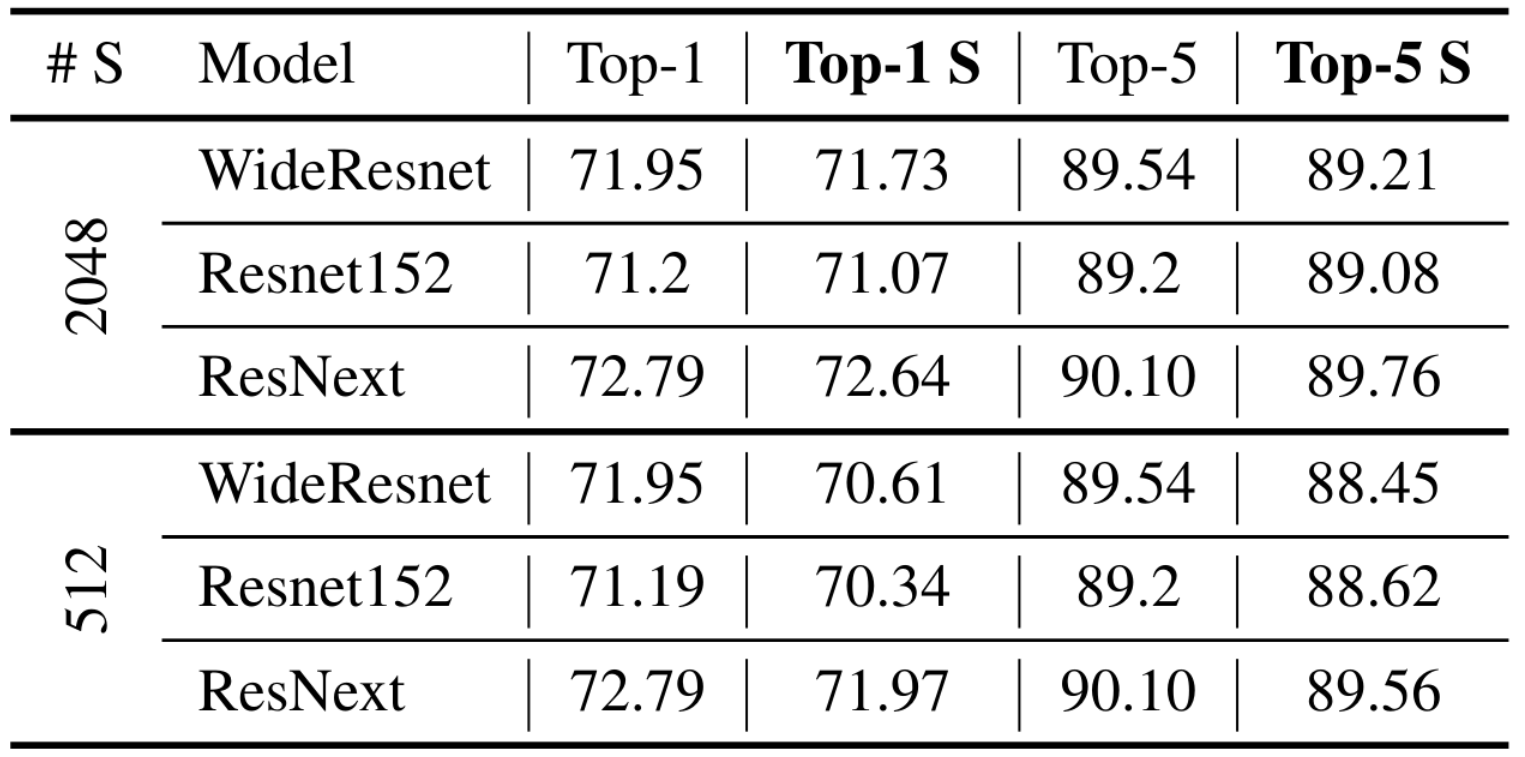}
}\label{tab:imagenetcleanacc}
\end{table}

For the other two hyperparameters, namely, the number of symbols representing all convolutional layer filters (N\_CFILTERS) and the number of symbols representing all fully connected layer filters (N\_FFILTERS), we perform a grid search to find the effect on accuracy for different choices of these (CIFAR-10 - VGG-11 combination). The results in Tab.~\ref{tab:hyper} show that the accuracy is primarily dependent on N\_CLUSTERS. 

For FPGA implementation, we used the configuration with N\_CLUSTERS=512,  N\_CFILTERS=256,  N\_FFILTERS=32.

\begin{table}[ht]
\caption{Grid search on symbol hyperparameters, namely, the number of symbols for representing input and activations (N\_CLUSTERS), the number of symbols representing all convolutional layer filters (N\_CFILTERS), and the number of symbols representing all fully connected layer filters (N\_FFILTERS) to find the effect on accuracy for different choices of these. Here we experiment on CIFAR-10 dataset on VGG-11 CNN} 
\centering
\scalebox{0.95}{
    \begin{tabular}{@{}cccc@{}}
        \toprule[.4mm]
        \textbf{N\_CLUST} & \textbf{N\_CFILT} & \textbf{N\_FFILT} & \textbf{ACC (\%)}  \\
        \midrule[0.4mm]
\textbf{512} & \textbf{256} & \textbf{128} & \textbf{89} \\\midrule[0.1mm]
\textbf{512} & \textbf{256} & \textbf{64} & \textbf{89} \\\midrule[0.1mm]
\textbf{512} & \textbf{256} & \textbf{32} & \textbf{89} \\\midrule[0.1mm]
\textbf{512} & \textbf{128} & \textbf{128} & \textbf{64} \\\midrule[0.1mm]
\textbf{512} & \textbf{128} & \textbf{64} & \textbf{64} \\\midrule[0.1mm]
\textbf{512} & \textbf{64} & \textbf{128} & \textbf{42} \\\midrule[0.1mm]
512 & 64 & 128 & 41 \\\midrule[0.1mm]
512 & 32 & 32 & 42 \\\midrule[0.1mm]
512 & 16 & 16 & 43 \\\midrule[0.1mm]
512 & 16 & 32 & 43 \\\midrule[0.1mm]
256 & 256 & 32 & 37 \\\midrule[0.1mm]
256 & 512 & 512 & 36 \\\midrule[0.1mm]
256 & 512 & 32 & 36 \\\midrule[0.1mm]
256 & 16 & 32 & 29 \\\midrule[0.1mm]
256 & 16 & 16 & 29 \\\midrule[0.1mm]
256 & 8 & 8 & 21 \\\midrule[0.1mm]
128 & 256 & 32 & 22 \\\midrule[0.1mm]
64 & 256 & 32 & 29 \\\midrule[0.1mm]
\bottomrule[0.4mm]
        \end{tabular}
\label{tab:hyper}
}
\end{table}

Based on the above design space exploration, we found that DietCNN always reduces the memory footprint of the model-dataset combinations that we used. Moreover, the negligible accuracy drop due to approximation, observed in Tab.~\ref{tab:imagenetcleanacc}, indicates that the DietCNN methodology has the potential to scale to much larger datasets that we have not attempted yet.

An illustrative example for the VGG-11 network is provided below:

\texttt{\{Image (32,32,3) -> Convolution (64,3,3,3) -> ReLU -> Pool (2), Stride (2)  -> Convolution (128,64,3,3), ->  ReLU  -> Pool (2), Stride (2)-> Convolution (256, 128,3,3), ->  ReLU  -> Convolution (256, 128,3,3), ->  ReLU -> Convolution (256, 256,3,3), ->  ReLU ->  Pool (2), Stride (2)-> Convolution (512, 256,3,3), ->  ReLU ->  Convolution (512, 512,3,3), ->  ReLU -> Pool (2), Stride (2) ->  ReLU ->  Convolution (512, 512,3,3), ->  ReLU ->  Convolution (512, 512,3,3), -> Pool (2), Stride (2) -> Linear \}}

The convolution kernels are shown in NCWH format. All convolutional layers have stride 1, not shown above. 

The DietCNN LUTs for this network are as follows:
\begin{enumerate}
    \item \textbf{Main LUTs:} 1. conv\_lut (80KB), 2. fc\_lut (92KB), 3. add\_lut (392KB), 4. relu\_lut (4KB), 5. centroid\_lut (4KB).
\item \textbf{Filter LUTs:} 6. c1\_sym\_filter (4KB), 7. c2\_sym\_filter (96KB), 8. c3\_sym\_filter (392KB), 9. c4\_sym\_filter (748KB), 10. c5\_sym\_filter (1.6MB), 11. c6\_sym\_filter (3.2MB), 12. c7\_sym\_filter (3.2MB), 13. c8\_sym\_filter (3.2MB), 14. f1\_sym\_filter (8KB)
\item \textbf{Bias LUTs:} 15. c1b\_lut (48KB), 16. c2b\_lut (88KB), 17. c3b\_lut (140KB), 18. c4b\_lut (140KB), 19. c5b\_lut (224KB), 20. c6b\_lut (256KB), 21. c7b\_lut (220KB), 22. c8b\_lut (204KB), 23. f1b\_lut (8KB)
\end{enumerate}

\textbf{Total:} 23 LUTs of size 14MB. Note that this is the final size of the DietVGG-11 model.

In contrast, the standard VGG-11 model has the following memory footprint:

\begin{enumerate}
    \item \textbf{Filters:}  1. c1f (8.0KB), 2. c2f (232KB), 3. c3f (892KB),  4. c4f (1.8MB), 5. c5f (3.3MB), 6. c6f (6MB), 7. c7f (5.3MB) 8. c8f (4.9MB) 9. f1f (16KB)
 \item \textbf{Biases:} 10. c1b (4KB) , 11. c2b (4KB), 12. c3b (4KB), 13. c4b (4KB), 14. c5b (4KB), 15. c6b (4KB),   16. c7b (4KB), 17. c8b (4KB), 18. c6b (4KB)   
\end{enumerate}
\textbf{Total:} Standard VGG-11 model 23MB  

The models with no bias result in a more drastic reduction in the model size, as shown in Section 4, Table 1, last column of the main paper for the ResNet-18 architecture.

Note that in an FPGA implementation, these weights are loaded as 32-bit floating-point values in the memory. This generates a much higher memory footprint compared to the DietCNN variant, for which the symbolic weights and biases are centroid/symbol numbers, loaded as ap\_uint7, that is, 7-bit unsigned integers.
		\begin{figure}[h]
		\begin{tabular}{m{0.5\hsize}}
		\centering
			\begin{subfigure}[t]{0.5\textwidth}
				\centering
				\includegraphics[width=0.89\textwidth]{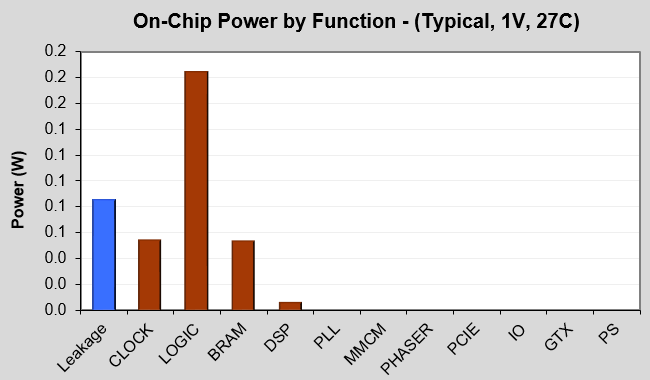}
				\caption{}\label{fig:forig1}
			\end{subfigure}
			
			\begin{subfigure}[t]{0.5\textwidth}
				\centering
				\includegraphics[width=0.89\textwidth]{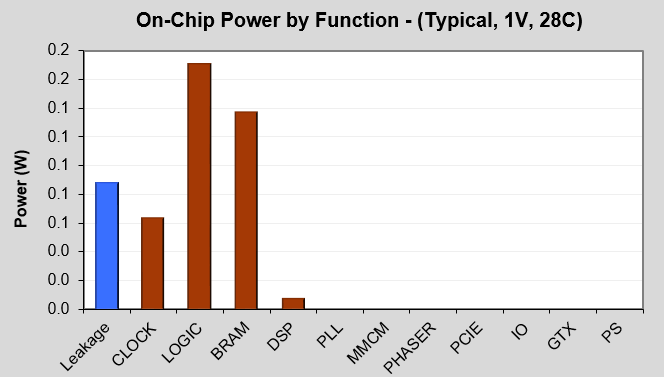}
				\caption{}\label{fig:forig2}
			\end{subfigure}
			
			\begin{subfigure}[t]{0.5\textwidth}
				\centering
				\includegraphics[width=0.89\textwidth]{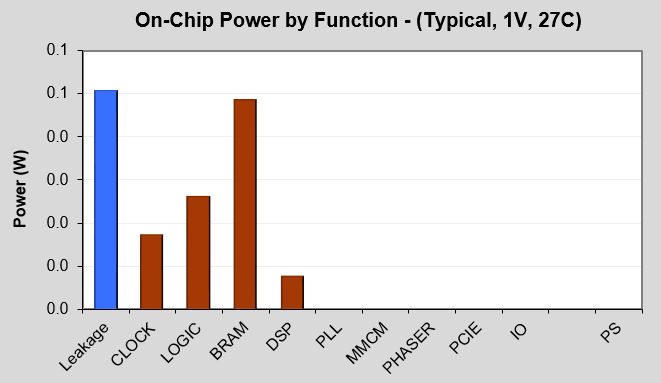}
				\caption{}\label{fig:forig3}
			\end{subfigure}

		\end{tabular}
		\caption{Power draw for the hardware components of DietCNN and two primary baselines, obtained from Xilinx XPE tool:
			\subref{fig:forig1} AdderNet power by functions;
			\subref{fig:forig2} ShiftAddNet power by functions;
			\subref{fig:forig3} DietCNN  power by functions;
		Lookup operations requires relatively less logic blocks (flip flop and LUTs) than norm and norm + shift.}\label{fig:funcpower}
	\end{figure}

	\begin{figure*}[!ht]
		\begin{tabular}{m{0.53\hsize}}
			\begin{subfigure}[t]{0.93\textwidth}
				\centering
				\includegraphics[width=0.99\textwidth]{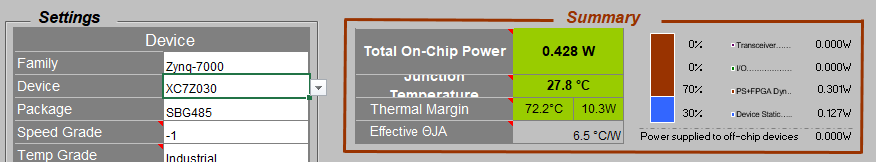}
				\caption{}\label{fig:morig1}
			\end{subfigure}
			
			\begin{subfigure}[t]{0.93\textwidth}
				\centering
				\includegraphics[width=0.99\textwidth]{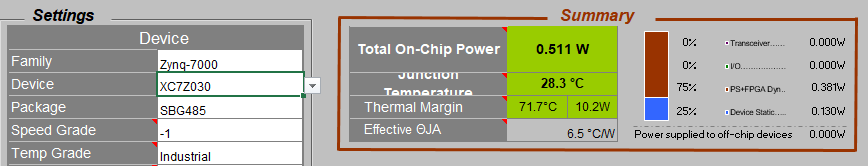}
				\caption{}\label{fig:morig2}
			\end{subfigure}
			
			\begin{subfigure}[t]{0.93\textwidth}
				\centering
				\includegraphics[width=0.99\textwidth]{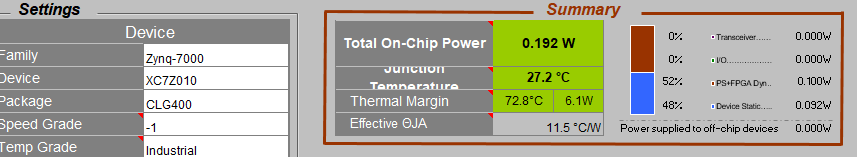}
				\caption{}\label{fig:morig3}
			\end{subfigure}

		\end{tabular}
		\caption{The main power draw for the DietCNN and two primary baselines, obtained from Xilinx XPE tool:
			\subref{fig:morig1} AdderNet intrinsic power;
			\subref{fig:morig2} ShiftAddNet intrinsic power;
			\subref{fig:morig3} DietCNN intrinsic power;
		}\label{fig:mainpower}
	\end{figure*}

\begin{table*}
\footnotesize\setlength{\tabcolsep}{3pt}
\caption{\small Illustrative example of VGG-11: calculation of the MAC operations reported in the Section 4, Table 1, column 3 of the main paper for the standard VGG-11 architecture. The pooling layers, non-linear activations and padding details are not shown.} 
\centering
\scalebox{0.98}{
  \begin{tabular}{@{}lcccccc cccccccc@{}}
  \toprule[.4mm]
    \textbf{Layer \#}  & 
    \multicolumn{2}{@{}c@{\hskip 4\tabcolsep}}{\textbf{Type }}  & 
    \multicolumn{2}{@{}c@{\hskip 4\tabcolsep}}{\textbf{Input}} &
    \multicolumn{2}{@{}c@{\hskip 4\tabcolsep}}{\textbf{Output}} &
    \multicolumn{2}{@{}c@{\hskip 4\tabcolsep}}{\textbf{Kernel}} &
    \multicolumn{2}{@{}c@{\hskip 4\tabcolsep}}{\textbf{Stride}} &
    \multicolumn{4}{@{}c@{\hskip 4\tabcolsep}}{\textbf{\# MAC}}  \\ 
    \midrule[0.4mm]
    1 & \multicolumn{2}{@{}c@{\hskip 4\tabcolsep}}{Convolution} & \multicolumn{2}{@{}c@{\hskip 4\tabcolsep}}{$32 \times 32\times3$} & \multicolumn{2}{@{}c@{\hskip 4\tabcolsep}}{$32\times32\times64$} & \multicolumn{2}{@{}c@{\hskip 4\tabcolsep}}{$3\times3\times64$} & \multicolumn{2}{@{}c@{\hskip 4\tabcolsep}}{$1\times1$} & \multicolumn{4}{@{}c@{\hskip 4\tabcolsep}}{$3 \times 3 \times 3 \times 32 \times 32 \times 64 = 1769472$}  \\
    \midrule[0.2mm]
    2 & \multicolumn{2}{@{}c@{\hskip 4\tabcolsep}}{Convolution} & \multicolumn{2}{@{}c@{\hskip 4\tabcolsep}}{$16\times16\times64$} & \multicolumn{2}{@{}c@{\hskip 4\tabcolsep}}{$16\times16\times128$} & \multicolumn{2}{@{}c@{\hskip 4\tabcolsep}}{$3\times3\times128$} & \multicolumn{2}{@{}c@{\hskip 4\tabcolsep}}{$1\times1$} & \multicolumn{4}{@{}c@{\hskip 4\tabcolsep}}{$3 \times 3 \times 64 \times 16 \times 16 \times 128 = 18874368$}  \\
    \midrule[0.2mm]
    3 & \multicolumn{2}{@{}c@{\hskip 4\tabcolsep}}{Convolution} & \multicolumn{2}{@{}c@{\hskip 4\tabcolsep}}{$8\times8\times128$} & \multicolumn{2}{@{}c@{\hskip 4\tabcolsep}}{$8\times8\times256$} & \multicolumn{2}{@{}c@{\hskip 4\tabcolsep}}{$3\times3\times256$} & \multicolumn{2}{@{}c@{\hskip 4\tabcolsep}}{$1\times1$} & \multicolumn{4}{@{}c@{\hskip 4\tabcolsep}}{$3 \times 3 \times 128 \times 8 \times 8 \times 256 = 18874368$}  \\
    \midrule[0.2mm]
    4 & \multicolumn{2}{@{}c@{\hskip 4\tabcolsep}}{Convolution} & \multicolumn{2}{@{}c@{\hskip 4\tabcolsep}}{$8\times8\times256$} & \multicolumn{2}{@{}c@{\hskip 4\tabcolsep}}{$8\times8\times256$} & \multicolumn{2}{@{}c@{\hskip 4\tabcolsep}}{$3\times3\times256$} & \multicolumn{2}{@{}c@{\hskip 4\tabcolsep}}{$1\times1$} & \multicolumn{4}{@{}c@{\hskip 4\tabcolsep}}{$3 \times 3 \times 256 \times 8 \times 8 \times 256 = 37748736$}  \\
    \midrule[0.2mm]
    5 & \multicolumn{2}{@{}c@{\hskip 4\tabcolsep}}{Convolution} & \multicolumn{2}{@{}c@{\hskip 4\tabcolsep}}{$4\times4\times256$} & \multicolumn{2}{@{}c@{\hskip 4\tabcolsep}}{$4\times4\times512$} & \multicolumn{2}{@{}c@{\hskip 4\tabcolsep}}{$3\times3\times512$} & \multicolumn{2}{@{}c@{\hskip 4\tabcolsep}}{$1\times1$} & \multicolumn{4}{@{}c@{\hskip 4\tabcolsep}}{$3 \times 3 \times 256 \times 4 \times 4 \times 512 = 18874368$}  \\
    \midrule[0.2mm]
    6 & \multicolumn{2}{@{}c@{\hskip 4\tabcolsep}}{Convolution} & \multicolumn{2}{@{}c@{\hskip 4\tabcolsep}}{$4\times4\times512$} & \multicolumn{2}{@{}c@{\hskip 4\tabcolsep}}{$4\times4\times512$} & \multicolumn{2}{@{}c@{\hskip 4\tabcolsep}}{$3\times3\times512$} & \multicolumn{2}{@{}c@{\hskip 4\tabcolsep}}{$1\times1$} & \multicolumn{4}{@{}c@{\hskip 4\tabcolsep}}{$3 \times 3 \times 512 \times 4 \times 4 \times 512 = 37748736$}  \\
    \midrule[0.2mm]
    7 & \multicolumn{2}{@{}c@{\hskip 4\tabcolsep}}{Convolution} & \multicolumn{2}{@{}c@{\hskip 4\tabcolsep}}{$2\times2\times512$} & \multicolumn{2}{@{}c@{\hskip 4\tabcolsep}}{$2\times2\times512$} & \multicolumn{2}{@{}c@{\hskip 4\tabcolsep}}{$3\times3\times512$} & \multicolumn{2}{@{}c@{\hskip 4\tabcolsep}}{$1\times1$} & \multicolumn{4}{@{}c@{\hskip 4\tabcolsep}}{$3 \times 3 \times 512 \times 2 \times 2 \times 512 = 9437184$}  \\
    \midrule[0.2mm]
    8 & \multicolumn{2}{@{}c@{\hskip 4\tabcolsep}}{Convolution} & \multicolumn{2}{@{}c@{\hskip 4\tabcolsep}}{$2\times2\times512$} & \multicolumn{2}{@{}c@{\hskip 4\tabcolsep}}{$2\times2\times512$} & \multicolumn{2}{@{}c@{\hskip 4\tabcolsep}}{$3\times3\times512$} & \multicolumn{2}{@{}c@{\hskip 4\tabcolsep}}{$1\times1$} & \multicolumn{4}{@{}c@{\hskip 4\tabcolsep}}{$3 \times 3 \times 512 \times 2 \times 2 \times 512 = 9437184$}  \\
    \midrule[0.2mm]
    9 & \multicolumn{2}{@{}c@{\hskip 4\tabcolsep}}{Linear} & \multicolumn{2}{@{}c@{\hskip 4\tabcolsep}}{$1\times512$} & \multicolumn{2}{@{}c@{\hskip 4\tabcolsep}}{$1\times10$} & \multicolumn{2}{@{}c@{\hskip 4\tabcolsep}}{$512\times10$} & \multicolumn{2}{@{}c@{\hskip 4\tabcolsep}}{$1$} & \multicolumn{4}{@{}c@{\hskip 4\tabcolsep}}{$512 \times 10 = 5120$}  \\
    \bottomrule[0.4mm]
    & \multicolumn{2}{@{}c@{\hskip 4\tabcolsep}}{} & \multicolumn{2}{@{}c@{\hskip 4\tabcolsep}}{} & \multicolumn{2}{@{}c@{\hskip 4\tabcolsep}}{\textbf{Total}} & \multicolumn{2}{@{}c@{\hskip 4\tabcolsep}}{} & \multicolumn{2}{@{}c@{\hskip 4\tabcolsep}}{} & \multicolumn{4}{@{}c@{\hskip 4\tabcolsep}}{$152769536 = 152.8$M}  \\
    \bottomrule[0.4mm]
  \end{tabular}
}
\label{tab:calcmac}
\end{table*}

\begin{table*}
\footnotesize\setlength{\tabcolsep}{3pt}
\caption{\small Illustrative example of VGG-11: calculation of the Lookup operations reported for the DietCNN variant of VGG-11 in the Section 4, Table 1, column 3 of the main paper. Compared to the standard VGG-11 in Tab.~\ref{tab:calcmac}, this DietCNN variant uses a stride of $2$ in the first convolutional layer and then onward no padding to ensure that the final output to the linear layer has the same feature map sizes. The final number of Lookups is the double of the total shown in the last row , one set for the multiplication LUT and another set for looking up the addition LUT} 
\centering
\scalebox{0.98}{
  \begin{tabular}{@{}lcccccc cccccccc@{}}
  \toprule[.4mm]
    \textbf{Layer \#}  & 
    \multicolumn{2}{@{}c@{\hskip 4\tabcolsep}}{\textbf{Type }}  & 
    \multicolumn{2}{@{}c@{\hskip 4\tabcolsep}}{\textbf{Input}} &
    \multicolumn{2}{@{}c@{\hskip 4\tabcolsep}}{\textbf{Output}} &
    \multicolumn{2}{@{}c@{\hskip 4\tabcolsep}}{\textbf{Kernel}} &
    \multicolumn{2}{@{}c@{\hskip 4\tabcolsep}}{\textbf{Stride}} &
    \multicolumn{4}{@{}c@{\hskip 4\tabcolsep}}{\textbf{\# Lookups}}  \\ 
    \midrule[0.4mm]
    1 & \multicolumn{2}{@{}c@{\hskip 4\tabcolsep}}{Convolution} & \multicolumn{2}{@{}c@{\hskip 4\tabcolsep}}{$32 \times 32\times3$} & \multicolumn{2}{@{}c@{\hskip 4\tabcolsep}}{$15\times15\times64$} & \multicolumn{2}{@{}c@{\hskip 4\tabcolsep}}{$3\times3\times64$} & \multicolumn{2}{@{}c@{\hskip 4\tabcolsep}}{$2\times2$} & \multicolumn{4}{@{}c@{\hskip 4\tabcolsep}}{$3\times3\times3\times15\times15\times64 = 388800$}  \\
    \midrule[0.2mm]
    2 & \multicolumn{2}{@{}c@{\hskip 4\tabcolsep}}{Convolution} & \multicolumn{2}{@{}c@{\hskip 4\tabcolsep}}{$15\times15\times64$} & \multicolumn{2}{@{}c@{\hskip 4\tabcolsep}}{$13\times13\times128$} & \multicolumn{2}{@{}c@{\hskip 4\tabcolsep}}{$3\times3\times128$} & \multicolumn{2}{@{}c@{\hskip 4\tabcolsep}}{$1\times1$} & \multicolumn{4}{@{}c@{\hskip 4\tabcolsep}}{$3\times3 \times 64 \times 13 \times 13 \times 128 = 12460032$}  \\
    \midrule[0.2mm]
    3 & \multicolumn{2}{@{}c@{\hskip 4\tabcolsep}}{Convolution} & \multicolumn{2}{@{}c@{\hskip 4\tabcolsep}}{$13\times13\times128$} & \multicolumn{2}{@{}c@{\hskip 4\tabcolsep}}{$11\times11\times256$} & \multicolumn{2}{@{}c@{\hskip 4\tabcolsep}}{$3\times3\times256$} & \multicolumn{2}{@{}c@{\hskip 4\tabcolsep}}{$1\times1$} & \multicolumn{4}{@{}c@{\hskip 4\tabcolsep}}{$3 \times 3 \times 128 \times 11 \times 11 \times 256 = 35684352$}  \\
    \midrule[0.2mm]
    4 & \multicolumn{2}{@{}c@{\hskip 4\tabcolsep}}{Convolution} & \multicolumn{2}{@{}c@{\hskip 4\tabcolsep}}{$11\times11\times256$} & \multicolumn{2}{@{}c@{\hskip 4\tabcolsep}}{$9\times9\times256$} & \multicolumn{2}{@{}c@{\hskip 4\tabcolsep}}{$3\times3\times256$} & \multicolumn{2}{@{}c@{\hskip 4\tabcolsep}}{$1\times1$} & \multicolumn{4}{@{}c@{\hskip 4\tabcolsep}}{$3 \times 3 \times 256 \times 9 \times 9 \times 256 = 47775744$}  \\
    \midrule[0.2mm]
    5 & \multicolumn{2}{@{}c@{\hskip 4\tabcolsep}}{Convolution} & \multicolumn{2}{@{}c@{\hskip 4\tabcolsep}}{$9\times9\times256$} & \multicolumn{2}{@{}c@{\hskip 4\tabcolsep}}{$7\times7\times512$} & \multicolumn{2}{@{}c@{\hskip 4\tabcolsep}}{$3\times3\times512$} & \multicolumn{2}{@{}c@{\hskip 4\tabcolsep}}{$1\times1$} & \multicolumn{4}{@{}c@{\hskip 4\tabcolsep}}{$3 \times 3 \times 256 \times 7 \times 7 \times 512 = 57802752$}  \\
    \midrule[0.2mm]
    6 & \multicolumn{2}{@{}c@{\hskip 4\tabcolsep}}{Convolution} & \multicolumn{2}{@{}c@{\hskip 4\tabcolsep}}{$7\times7\times512$} & \multicolumn{2}{@{}c@{\hskip 4\tabcolsep}}{$5\times5\times512$} & \multicolumn{2}{@{}c@{\hskip 4\tabcolsep}}{$3\times3\times512$} & \multicolumn{2}{@{}c@{\hskip 4\tabcolsep}}{$1\times1$} & \multicolumn{4}{@{}c@{\hskip 4\tabcolsep}}{$3 \times 3 \times 512 \times 5 \times 5 \times 512 = 58982400$}  \\
    \midrule[0.2mm]
    7 & \multicolumn{2}{@{}c@{\hskip 4\tabcolsep}}{Convolution} & \multicolumn{2}{@{}c@{\hskip 4\tabcolsep}}{$5\times5\times512$} & \multicolumn{2}{@{}c@{\hskip 4\tabcolsep}}{$3\times3\times512$} & \multicolumn{2}{@{}c@{\hskip 4\tabcolsep}}{$3\times3\times512$} & \multicolumn{2}{@{}c@{\hskip 4\tabcolsep}}{$1\times1$} & \multicolumn{4}{@{}c@{\hskip 4\tabcolsep}}{$3 \times 3 \times 512 \times 3 \times 3 \times 512 = 21233664$}  \\
    \midrule[0.2mm]
    8 & \multicolumn{2}{@{}c@{\hskip 4\tabcolsep}}{Convolution} & \multicolumn{2}{@{}c@{\hskip 4\tabcolsep}}{$3\times3\times512$} & \multicolumn{2}{@{}c@{\hskip 4\tabcolsep}}{$1\times1\times512$} & \multicolumn{2}{@{}c@{\hskip 4\tabcolsep}}{$3\times3\times512$} & \multicolumn{2}{@{}c@{\hskip 4\tabcolsep}}{$1\times1$} & \multicolumn{4}{@{}c@{\hskip 4\tabcolsep}}{$3 \times 3 \times 512 \times 1 \times 1 \times 512 = 2359296$}  \\
    \midrule[0.2mm]
    9 & \multicolumn{2}{@{}c@{\hskip 4\tabcolsep}}{Linear} & \multicolumn{2}{@{}c@{\hskip 4\tabcolsep}}{$1\times512$} & \multicolumn{2}{@{}c@{\hskip 4\tabcolsep}}{$1\times10$} & \multicolumn{2}{@{}c@{\hskip 4\tabcolsep}}{$512\times10$} & \multicolumn{2}{@{}c@{\hskip 4\tabcolsep}}{$1$} & \multicolumn{4}{@{}c@{\hskip 4\tabcolsep}}{$512 \times 10 = 5120$}  \\
    \bottomrule[0.4mm]
    & \multicolumn{2}{@{}c@{\hskip 4\tabcolsep}}{} & \multicolumn{2}{@{}c@{\hskip 4\tabcolsep}}{} & \multicolumn{2}{@{}c@{\hskip 4\tabcolsep}}{\textbf{Total}} & \multicolumn{2}{@{}c@{\hskip 4\tabcolsep}}{} & \multicolumn{2}{@{}c@{\hskip 4\tabcolsep}}{} & \multicolumn{4}{@{}c@{\hskip 4\tabcolsep}}{$236692160 = 236.6$M}  \\
    \bottomrule[0.4mm]
  \end{tabular}
}
\label{tab:symcalcmac}
\end{table*}

\section{Details of DietCNN Energy Measurement}
\label{sec:fpga}
This part of the appendix supplements Sec.IV of the main paper.\\

\noindent \textbf{Power Estimation Experiments}
This subsection reports the details of power estimation experiments, continuing from Sec. 4.3 of the main paper.  
Power measurement done here is an estimate obtained using Xilinx Power Estimator (XPE)\cite{xpe}. 
Using this tool we estimate the intrinsic power (Watt) of the CNN architecture on board, and then calculate the energy used (Joules) using the power and latency per inference (E(J) = P(W) $\times$ t(s)). For FPGA inference we use a batch size of 1. 
The Energy calculation steps are as follows:
\begin{itemize}
    \item We performed C-Synthesis and RTL co-simulation on XILINX VITIS HLS. We download the VIVADO IP and measure both static power and the power draw due to the CNN model.  
\item We measure the latency per inference with a batch size of 1
\item We set the clock period as 10 nanoseconds (100Mhz) 
\item We obtain the power from XPE for running the CNN variant on board in Watt
\item We calculate Watts to joules: The energy (E) is equal to the power (P) in times the time period t in seconds (s), which is the latency for one inference.
\end{itemize}
Fig.~\ref{fig:mainpower} and Fig.~\ref{fig:funcpower} show the total power on the board and power by functions respectively, for DietCNN and two comparison baselines AdderNet~\cite{chen2020addernet} and ShiftAddNet~\cite{hao2020shift}.\\

\noindent \textbf{Calculation of Operations in the Models}
This subsection presents the methodology followed for counting the number of Multiply-and-Accumulate (MAC) operations in the standard CNNs and the lookup operations in the corresponding DietCNN variants. These values were reported in the Table 1, column 3 of the main paper.

The formula for counting the operations in the convolutional layer is as follows:
 \begin{equation} 
 \nonumber
 \begin{split}
(\, filter\_height \,\times\, filter\_width \,\times\, input\_channels \\
\,\times\, output\_size \,\times\, output\_width \,\times\, output\_channels)
\end{split}
\end{equation}

The total number of MAC operations shown in Tab.~\ref{tab:calcmac} is for the standard VGG-11. In AdderNets, these MACs are replaced by equal number of L1 norms. In ShiftAddNets, one MAC is replaced by one L1 norm and one shift operation. Both these networks require addition of some batch normalization layers for their training to converge properly.  \par In contrast, Tab.~\ref{tab:symcalcmac} shows the number of Lookups that is required for replacing the Multiplications in the DietCNN variant. Using this methodology we measure and report the number of operations for the other models in Table 1, column 3 of the main paper.

\section{Details of Assets Used}\label{sec:assets}
This part provides additional details over Sec. 3.1 of the main paper.

\noindent \textbf{Clustering \& Dictionary Learning}
We use FAISS~\cite{johnson2021faiss} to cluster the images and intermediate feature maps. We use Scikit-learn~\cite{scikit-learn} K-means++ algorithm to cluster the weights and biases.

\noindent \textbf{CNN Training}
We use PyTorch~\cite{pytorch2019} to implement the DietCNN retraining.  

\noindent \textbf{Experimentation Platform}
$32$ core Intel(R) Xeon(R) Silver $4108$ CPU \@ $1.80$GHz workstation, running Ubuntu 20.04 with $128$ Gibibytes of RAM, and NVIDIA P1000 GPU (4 Gigabytes memory). 

\noindent \textbf{FPGA C Synthesis}
FPGA implementations were done on Zynq 7000 boards using the Xilinx Vitis tool and power estimations were carried out using Xilinx Power Estimator (XPE)~\cite{xpe}.

\end{document}